%% file: root.tex
\documentclass[letterpaper, 10 pt, journal, twoside]{IEEEtran}
\usepackage[utf8]{inputenc}
\usepackage{url}
\usepackage{cite}
\usepackage{amsmath,amssymb,amsfonts}
\usepackage{textcomp}
\usepackage{lettrine}
\usepackage{caption}
\usepackage{subcaption}
\usepackage[export]{adjustbox}
\newcommand{\subparagraph}{}
\usepackage{leftidx}
\usepackage{siunitx}
\usepackage{multirow}
\usepackage{balance}
\usepackage{booktabs}
\DeclareMathOperator*{\argmin}{arg\,min}

\usepackage{hyperref}
\usepackage{algorithm2e}
\RestyleAlgo{ruled}
\usepackage{mathtools}
\usepackage{setspace}
\usepackage{flushend}
\usepackage{titlesec}
\titlespacing{\section}{0pt}{*1}{*0.5}
\titlespacing{\subsection}{0pt}{*0.8}{*0.4}
\usepackage{setspace}
\begin{document}

%\title{\LARGE \bf mS-Graphs: A Real-time and Efficient Localization and Mapping for Large Multi-floor Indoor Environments}

\title{\LARGE \bf {S-Graphs 2.0} -- A Hierarchical-Semantic Optimization and Loop Closure for SLAM}

\author{Hriday Bavle$^{1}$, Jose Luis Sanchez-Lopez$^{1}$, Muhammad Shaheer$^{1}$, \\ Javier Civera$^{2}$ and Holger Voos$^{1}$ % <-this % stops a space
\thanks{Manuscript received: February, 28, 2025; Accepted: September, 19, 2025.}
\thanks{This paper was recommended for publication by Editor Sven Behnke upon evaluation of the Associate Editor and Reviewers’ comments.}
\thanks{*This research was funded in whole, or in part, by the Luxembourg National Research Fund (FNR), DEUS Project, ref. C22/IS/17387634/DEUS. For the purpose of open access, and in fulfilment of the obligations arising from the grant agreement, the author has applied a Creative Commons Attribution 4.0 International (CC BY 4.0) license to any  Author Accepted Manuscript version arising from this submission.}%
\thanks{$^{1}$Authors are with the Automation and Robotics Research Group, Interdisciplinary Centre for Security, Reliability and Trust, University of Luxembourg. Holger Voos is also associated with the Faculty of Science, Technology and Medicine, University of Luxembourg, Luxembourg.
\tt{\small{\{hriday.bavle, joseluis.sanchezlopez, muhammad.shaheer, holger.voos\}}@uni.lu}}% 
\thanks{$^{2}$Author is with I3A, Universidad de Zaragoza, Spain
{\tt\small jcivera@unizar.es}}%
\thanks{Digital Object Identifier (DOI): see top of this page.}
}

\markboth{IEEE Robotics and Automation Letters. Preprint Version. Accepted September, 2025}
{Bavle \MakeLowercase{\textit{et al.}}: \textit{S-Graphs 2.0}} 

\maketitle
% \thispagestyle{empty}
% \pagestyle{empty}
% Comment or remove these lines for final RAL version.
\input{abstract}
\input{introduction}
\input{related_works}
\input{system_overview}
\input{experimental_results}
\input{conclusion}
\bibliographystyle{IEEEtran}
\bibliography{Bibliography}
%\balance
%\vfill
%\clearpage
%\appendix
%\input{appendix}
\end{document}

%% file: abstract.tex
\begin{abstract}
The hierarchical nature of 3D scene graphs aligns well with the structure of man-made environments, making them highly suitable for representation purposes. Beyond this, however, their embedded semantics and geometry could also be leveraged to improve the efficiency of map and pose optimization, an opportunity that has been largely overlooked by existing methods.
We introduce Situational Graphs 2.0 (\textit{S-Graphs 2.0}), that effectively uses the hierarchical structure of indoor scenes for efficient data management and optimization.  Our approach builds a four-layer situational graph comprising \textit{Keyframes}, \textit{Walls}, \textit{Rooms}, and \textit{Floors}. Our first contribution lies in the front-end, which includes a floor detection module capable of identifying stairways and assigning floor-level semantic relations to the underlying layers (\textit{Keyframes}, \textit{Walls}, and \textit{Rooms}). Floor-level semantics allows us to propose a floor-based loop closure strategy, that effectively rejects false positive closures that typically appear due to aliasing between different floors of a building. Our second novelty lies in leveraging our representation hierarchy in the optimization. Our proposal consists of: (1) local optimization over a window of recent keyframes and their connected components across the four representation layers, (2) floor-level global optimization, which focuses only on keyframes and their connections within the current floor during loop closures, and (3) room-level local optimization, marginalizing redundant keyframes that share observations within the room, which reduces the computational footprint. We validate our algorithm extensively in different real multi-floor environments. Our approach shows state-of-the-art accuracy metrics in large-scale multi-floor environments, estimating hierarchical representations up to $10\times$ faster, in average, than competing baselines.
Our code is open-sourced at: \mbox{\url{https://github.com/snt-arg/lidar_situational_graphs}}
\end{abstract}

%% file: introduction.tex
\section{Introduction}
Simultaneous Localization and Mapping (SLAM), that has typically focused on geometric aspects, should also enable a robot with a high-level understanding of its environment \cite{cadena2016past}. The combination of 3D scene graphs with traditional SLAM graphs have recently sparked great interest given its compact and intuitive approach to representing an environment as a hierarchy of layers \cite{hydra}, \cite{curb_sg}. In this direction, Situational Graphs (\textit{S-Graphs}) \cite{s_graphs+} model the environment as a four-layered optimizable graph composed of of \textit{Keyframes}, \textit{Walls}, \textit{Rooms} and \textit{Floors}. As the robot explores large environments, however, the size of the graph grows, along with the its computational footprint. Essentially, while \textit{S-Graphs} and 3D scene graphs are motivated by the need of higher-level representations, none of them have yet tailored their optimization to the specifics of these hierarchical representations. Leveraging them will result in a lower computational footprint, enabling real-time graph optimization for larger maps, which will result finally in enhanced accuracy.
\begin{figure}[t]
  \centering
  \includegraphics[width=0.8\columnwidth]{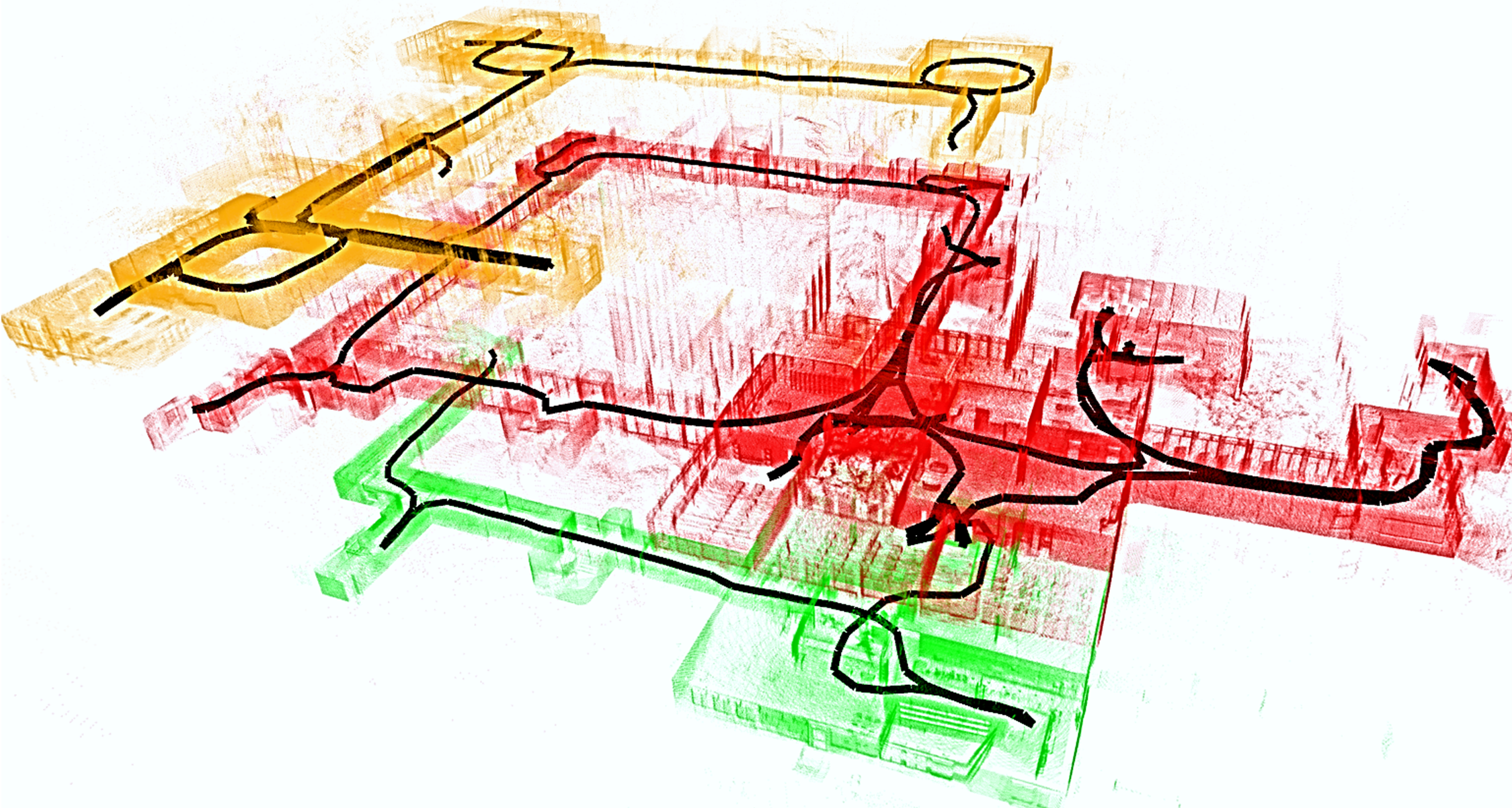}
  \caption{Three-story map generated by our \textit{S-Graphs 2.0}. Floor levels are color-coded in green, red, and orange, while the estimated robot trajectory is shown in black.}
  \label{fig:front_image}
\end{figure}
In this paper we present Situational Graphs 2.0 (\textit{S-Graphs 2.0}), that builds over the work of \cite{s_graphs+} and contributes with two novel approaches that improves its robustness and computational footprint. Our first novelty consists on proposing a floor detection module capable of segmenting different floor levels and the stairways that connect them. This floor-level information is incorporated in all the rooms, walls and keyframes layers of the graph ensuring robust localization and loop closures, even across similarly looking areas on different floors. Our second novelty is the proposal of an optimization strategy that leverages the hierarchical structure of the graph to reduce its computational complexity, while maintaining pose and map accuracy. Specifically, as a robot explores an environment at a given floor-level, it estimates a hierarchical graph by local optimization over a subset of the last $n$ keyframes and its connected layers. For loop closures at a given floor-level, we run a floor-level global optimization, optimizing only the subgraph of that floor-level. Additionally, on detection of a room candidate bounded by four walls, we exploit the \textit{room-wall} hierarchy to perform room-level local optimization to marginalize keyframes and its connections removing redundant keyframes observing the same room. To summarize our main contributions are:
\begin{itemize}
    \item \textit{S-Graphs 2.0}, a hierarchical SLAM leveraging semantic relations for efficient management and optimization of robot poses and map elements.
    \item Floor-level loop closure for robust localization in similar-looking environments at different levels. 
    \item Floor- and room-level hierarchical optimization strategies, marginalizing out unnecessary entities while maintaining the map accuracy.
\end{itemize}
\begin{figure*}[!h]
  \centering
  \includegraphics[width=0.95\textwidth]{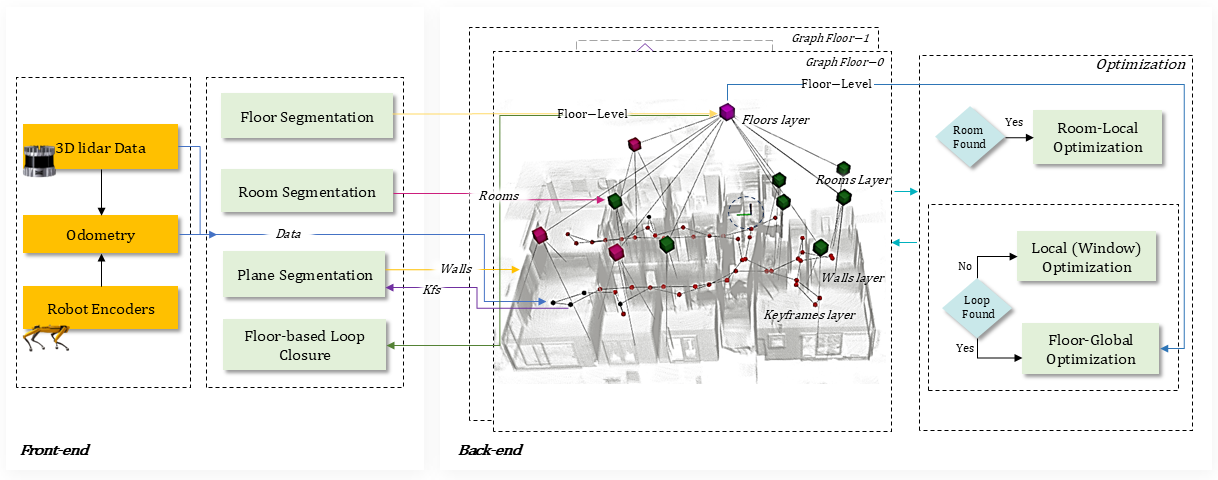}
  \caption{\textbf{System Architecture.} The inputs to our method are the 3D LiDAR data along with the odometry measurements. Its contains different modules in the front-end modules for generating the four-layered hierarchical graph and organizing it into floor-level. A back-end module which exploits the hierarchy in the graph to apply different optimization strategies.}
  \label{fig:system_architecture}
  \vspace{-0.3cm}
\end{figure*}

%% file: related_works.tex
\section{Related Works}
\label{sec:related_works}
\subsection{Simultaneous Localization and Mapping}
\textbf{Metric-Semantic SLAM.} The most referenced LiDAR SLAM pipelines include LOAM \cite{loam}, LIO-SAM \cite{lio_sam}, BALM \cite{BALM}, Fast-LIO \cite{fast_lio}, \cite{fast_lio2}. All the above techniques utilize low-level geometric features in the environment to estimate the pose and map of the robot, due to which such techniques can be limited in their accuracy when exploring large and complex indoor environments. Additionally several other SLAM techniques exist in the literature that utilize high-level semantic features additional to geometric features to improve the environment understanding and the map and pose accuracy. Some of such techniques include LeGO-LOAM \cite{lego-loam}, SA-LOAM \cite{sa-loam}, SegMap \cite{segmap}, SUMA++ \cite{suma}. LeGO-LOAM \cite{lego-loam} utilizes different planar semantics in the environment like ground plane to improve the map and pose estimate. While SA-LOAM \cite{sa-loam} utilizes high-level semantics to improve the loop closure accuracy of the underlying metric SLAM, SegMap \cite{segmap} utilizes learned high-level descriptors from the environment to perform robust localization with respect to the high-level descriptors. SUMA++ \cite{suma} completely segments the environment in different semantic features to perform object level semantic SLAM, removing dynamic entities from the scene. 

However, one of the major limitations of the metric-semantic SLAM techniques is that they do no exploit the different semantic entities to perform a better map management strategies for improved optimization and map accuracy for large scale environments. Most of these techniques either clear major map elements as the map size increases to maintain real-time performance or do not provide a real-time analysis of the underlying SLAM when managing large scale maps.    

\textbf{Hierarchical SLAM.} 
To overcome the inherent problem of improving the computation with increasing map size, works like \cite{hierarchical_optimization}, \cite{information_theoretic} exploit the methods to compress the graph into different sub-graphs to maintain real-time performance without the loss of map and pose accuracy.  \cite{hierarchical_optimization} present a technique grouping nodes into different sub-graphs based on a simple distance based criteria. While \cite{information_theoretic} present an information-theoretic approach for factor graph compression where laser scans measurements
and their corresponding robot poses are removed such that
the expected loss of information with respect to the current
map is minimized. In \cite{continous_time_slam} authors present a hierarchical continuous time SLAM algorithm dividing it into local sub-graphs aggregating measurements from 3D LiDAR, the generated sub-graphs are aligned with each other using local surfel-based registration techniques. In the above technique authors also use distance-based heuristic to create different sub-graphs for the optimization problem. \cite{globally_consistent} propose a local and global hierarchical optimization technique using sub-map strategies similar to \cite{cartographer}. They generate local sub-maps at given distance intervals performing local optimization and then connecting different local sub-maps using global registration to perform efficient global optimization. The authors in \cite{globally_consistent} also choose a heuristic based stragety to generate the local and global maps. 

Although hierarchical SLAM shows efficient optimization of the graph for real-time performance, currently most of the techniques rely on time-distance based heuristics to formulate and optimize the hierarchical graph. 
\subsection{3D Scene Graphs}
Recently, 3D scene graphs have shown great potential in representing the environment in a more meaningful and compact manner \cite{armeni, 3dssg, 3dscene_graph, scene_graph_fusion, 3ddsg}. Additionally methods such as, \cite{hydra}, \cite{curb_sg}, \cite{s_graphs+} tightly couple the 3D scene graph with SLAM graphs exploiting the hierarchy in the environment to improve the final pose and map accuracy. \cite{hydra} focuses on the generation of a 3D scene graph mainly using RGB-D cameras, incorporating different hierarchies of objects, places, rooms and buildings, furthermore it utilizes the hierarchy to perform a top-down and bottom-up search improving the search of loop closure candidates, thus improving the final map accuracy. \cite{curb_sg} presents a 3D scene graph for the outdoor environment, dividing it into hierarchies such as lanes, landmarks, intersections, and environment. They factor different connections of the hierarchical graph as a factor graph, continuously optimizing the pose and map. \cite{situational_graphs, s_graphs+} represent an indoor scene as a four-layered optimizable factor graph dividing it into layers of keyframes, walls, rooms, and floors, simultaneously optimizing all the layers to obtain improved map and pose accuracy. 

However, most of the works based on 3D scene graphs only focus on generating the hierarchical representation of the environment and do not exploit the intuitive nature of these hierarchical graphs to perform enhanced management/optimization of the different map elements which eventually leads to the problem of increased computation with the increase in the size of the explored scene. 

Thus in this work, we apply the concept of hierarchical SLAM to 3D scene graphs. The 3D scene graph structure allows to generate a meaningful hierarchical SLAM graph instead of using distance-time based heuristics. This hierarchical SLAM graph improves the management of generated map elements to scale for large scale environments, while maintaining the pose and map accuracy with real-time performance.

%% file: system_overview.tex
\section{Overview}\label{overview}
\subsection{Background}
\textbf{Situational Graphs.} We build on S-Graphs+ \cite{s_graphs+}, that estimates a layered graph representation of the environment from 3D LiDAR. The four layers of the graph, illustrated in Fig.~\ref{fig:system_architecture}, are as follows. The \textbf{\textit{Keyframes Layer}} contains robot poses at different distance-time intervals. The \textbf{\textit{Walls Layer}} represents the walls in the environment as planar surfaces connected to the keyframes from which they were observed. The \textbf{\textit{Rooms Layer}} models two-wall and four-wall rooms, connecting in this manner the previous walls layer. Finally, the \textbf{\textit{Floors Layer}} represents the different floors in the building, connecting them with the underlying rooms layer.  
\subsection{System Architecture} \label{subsec:system_architecture}
Fig.~\ref{fig:system_architecture} shows the complete architecture of our approach.The front-end implements plane segmentation as in \cite{s_graphs+} and room segmentation as in \cite{millan2024learning}, with the addition of a floor segmentation module able to detect changes in the floor levels by measuring changes in the slope of the trajectory of keyframes. In practice, floor segmentation will be highly relevant to perform loop closure only within the same floor. This is a reasonable strategy, as local sensor readings are commonly limited to a single floor and connecting staircases. Additionally, it removes a high rate of false positives, caused by aliasing between floor levels.

Our back-end of the proposed approach acts on a four-layered optimizable factor graph, as in \cite{s_graphs+}, but adds several optimization strategies that leverage the hierarchy in the graph. Specifically, the optimization is performed at three levels. A local optimization acts over a window of $n$ keyframes every time new nodes or edges are added to the factor graph. A floor-level global optimization runs every time a loop closure candidate is found between two keyframes. Unlike typical global optimization threads in SLAM, that operate on all keyframes, our floor-level optimization focuses on a subset of the graph containing only the current floor level. It optimizes the keyframes and all the connected layers in the current floor level, and previously connected floor levels are only optimized when the current floor is revisited. A room-level local optimization marginalizes redundant keyframes and its connections observing the same room. This hierarchical optimizations minimizes computational overhead while maintaining the map and pose accuracy across connected floor levels. The global state $\mathbf{s}$ that we estimate is as follows
\begin{multline}
\mathbf{s} = [
\leftidx{^M}{\boldsymbol{\xi}}_{{1}}, \ \hdots, \ \leftidx{^M}{\boldsymbol{\xi}}_{F}, \\ 
\leftidx{^M}{\mathbf{x}}_{{R_1}_{\boldsymbol{\xi}_1}}, \ \hdots, \ \leftidx{^M}{\mathbf{x}}_{{R_T}_{\boldsymbol{\xi}_F}}, \ \leftidx{^M}{\boldsymbol{\pi}}_{{1}_{\boldsymbol{\xi}_1}}, \ \hdots, \ \leftidx{^M}{\boldsymbol{\pi}}_{{P}_{\boldsymbol{\xi}_F}}, \\
\leftidx{^M}{\boldsymbol{\rho}}_{{1}_{\boldsymbol{\xi}_1}}, \ \hdots, \ \leftidx{^M}{\boldsymbol{\rho}}_{{S}_{\boldsymbol{\xi}_F}}, \ \leftidx{^M}{\boldsymbol{\kappa}}_{{1}_{\boldsymbol{\xi}_1}}, \ \hdots, \  \leftidx{^M}{\boldsymbol{\kappa}}_{{K}_{\boldsymbol{\xi}_F}}, \ \\  \ \leftidx{^M}{\mathbf{x}}_{O}]^\top
\end{multline}
where $\leftidx{^M}{\boldsymbol{\xi}}_{f} \in \mathbb{R}^3, f \in \{1, \hdots , \ F \}$ are the $F$ floors levels. All the nodes in the state at a given time contain the semantic of the floor-level it belongs to. $\leftidx{^M}{\mathbf{x}}_{{R_t}_{\boldsymbol{\xi}_f}} \in SE(3), \ t \in \{1, \hdots, T\}$ are the robot poses belonging to a given floor level. $\leftidx{^M}{\boldsymbol{\pi}}_{{i}_{\boldsymbol{\xi}_f}} \in \mathbb{R}^3, \ i \in \{1, \hdots, P\}$ are the plane parameters of the $P$ wall planes belonging to a given floor level. $\leftidx{^M}{\boldsymbol{\rho}}_{{j}_{\boldsymbol{\xi}_f}} \in \mathbb{R}^3, \ j \in \{1, \hdots, S\}$ contains the parameters of the $S$ four-wall rooms and $\leftidx{^M}{\boldsymbol{\kappa}}_{k} \in \mathbb{R}^3, \ k \in \{1, \hdots, K\}$ the parameters of the $K$ two-wall rooms, all belonging to a given floor-level ${\boldsymbol{\xi}_f}$. Finally, $\leftidx{^M}{\mathbf{x}}_{O}$ is the estimated drift between the map frame and the odometry frame. 
\begin{figure}[b]
  \centering
  \includegraphics[width=0.4\columnwidth]{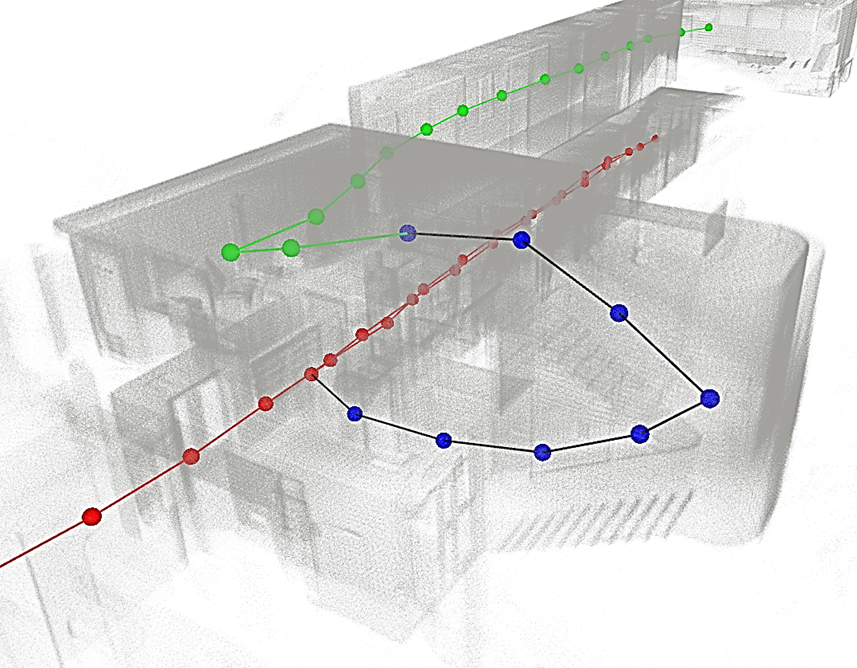}
  \caption{\textbf{Stairway Detection}, with the keyframes corresponding to stairways as blue dots, and those corresponding to different floor levels as red and green dots.}
  \label{fig:stairway_detection}
\end{figure}
\section{Front-End}
\subsection{Floor Segmentation}
\textbf{Floor Center.} The floor segmentation module extracts the center of a current floor level from the largest wall planes currently extracted in a scene. Each time the pipeline is run, it creates a default floor node with the center placed at the origin of frame $M$. It then utilizes the information from all walls at the current floor level $^M\boldsymbol{\xi}_f$ to create a sub-category of wall planes $\leftidx{^M}{\boldsymbol{\Pi}_{i_{\boldsymbol{\xi}_f}}}$ ($^M\boldsymbol{\Pi}$ represents a set of $^M\pi$ planes) where $i=\{1, \hdots, T \}$. The walls sub-categories consist of $x$ vertical planar surfaces and $y$ vertical planar surfaces based on their normal orientation. After receiving a plane set, it computes the widths ${w}_x$ between all $x$-direction planes and similarly ${w}_y$ for $y$-direction planes as in \cite{s_graphs+}. 

The wall plane set with the largest $w_x$ and $w_y$ is the chosen candidate for the current floor level. These planar pairs in both $x$ and $y$ direction undergo an additional dot product check between their corresponding normal orientations, $|\mathbf{n}_{x_{a_1}} \cdot \mathbf{n}_{x_{b_1}}| < t_n$ and $|\mathbf{n}_{y_{a_1}} \cdot \mathbf{n}_{y_{b_1}}| < t_n$, to remove wall planes originating outside the building structure. The floor segmentation then computes the floor center node using the obtained wall plane candidates as explained in \cite{s_graphs+}.  

\textbf{Stairway Detection.} 
The stairway detection module is responsible for identifying transitions between floor levels and associating corresponding keyframes with the detected stairs. We follow a process similar to \cite{lexis} but extend it to our use case for detecting multiple floors and associating keyframes belonging to stairways and different floor-levels (see Fig.~\ref{fig:stairway_detection}). 

Our process relies on the analysis of sequential keyframes captured during navigation, particularly focusing on their vertical displacement. 
It begins by maintaining a queue of keyframes that are sequentially analyzed to compute the slope of their vertical trajectory. The slope is calculated using linear regression on the height values extracted from the poses of the keyframes. This slope quantifies the rate of vertical displacement and serves as the primary indicator for stairway traversal with its corresponding sign distinguishing upward or downward movements. A slope above threshold $\tau_s$ is identified as the start of the stairway, after which all the keyframes are added into stairway keyframe sequence. When the gradient drops below $\tau_s$ we mark end of that stairway, with all the keyframes $k_s$ within the sequence being part of the stairway sequence. A new floor node is then created with its height set using the height of the final keyframe in the sequence and all subsequent keyframes assigned to the new floor-level. 
\subsection{Floor-based Loop Closure}
One of the challenges for SLAM in large environments with multiple floors is the similarity in the geometry and visual appearance of rooms and environments in same areas of the different floors. This aliasing cause conventional geometry- or semantic-based loop closure algorithms to fail, leading to incorrect associations between keyframes from different floors. Such errors can severely degrade the accuracy of the estimated maps.

To address this issue, we leverage the inherent hierarchical structure of our graph in combination with the floor segmentation module to incorporate floor-level information across all layers. Specifically, given a complete set of keyframes $^M\boldsymbol{K}_i$ we create a subset of keyframes $^M\boldsymbol{K}_f$, $\forall f \in ^M\boldsymbol{\xi}_f$ belonging to the current floor-level $ ^M\boldsymbol{\xi}_f$, and perform a scan matching-based loop-closure algorithm \cite{s_graphs+} for this subset, ensuring that keyframes from different floors are excluded from matching. 

To avoid false positives in transitional scenarios, such as climbing up/down of the stairs, our floor-based loop closure is temporarily disabled in such areas, as LiDAR measurements might capture regions that may not distinctly belong to any single floor. Floor-based loop closure is resumed once there is a full transition to a new floor-level. 
\subsection{Room Keyframe Segmentation} \label{sec:room_keyframe_segmentation}
In order to perform room-level local optimization (explained in Section \ref{sec:back_end}) we need to identify keyframes which are bounded within a four-wall room. We exploit the hierarchy within a room and its corresponding four-walls to identify the keyframes set $^M\boldsymbol{K}_{\rho_{i}}$ belonging to a room $^M\rho_i$. To compute if a keyframe lies within a room, we first compute the vector $\boldsymbol{v}_d = \boldsymbol{p} - \boldsymbol{q}_i$, where $\boldsymbol{p}$ is the translation component of the keyframe and
% $\boldsymbol{q}_i$ is the point on one of the planes belonging to the room. 
 $\boldsymbol{q}_i$ is a randomly selected point from the $i$-th plane of the room, where $i \in \{1, 2, 3, 4\}$ corresponds to each of the four planes belonging to the room.
 We then evaluate the dot product of $\delta = \boldsymbol{n}_i \cdot \boldsymbol{v}_d$, where $\boldsymbol{n}_i$ is the normal orientation of the plane. Keyframe positions with $\delta$ positive are considered to be bounded within the walls of the room. 
\section{Back-End} \label{sec:back_end}
The global state of the robot defined in Section \ref{subsec:system_architecture} is optimized in three different stages exploiting the hierarchy of the graph. We mainly detail the optimization strategies, where relies the novelty of the paper. Details on the formulation of the cost function can be found in \cite{s_graphs+}.
\begin{figure}[b]
  \centering
  \includegraphics[width=.8\columnwidth]{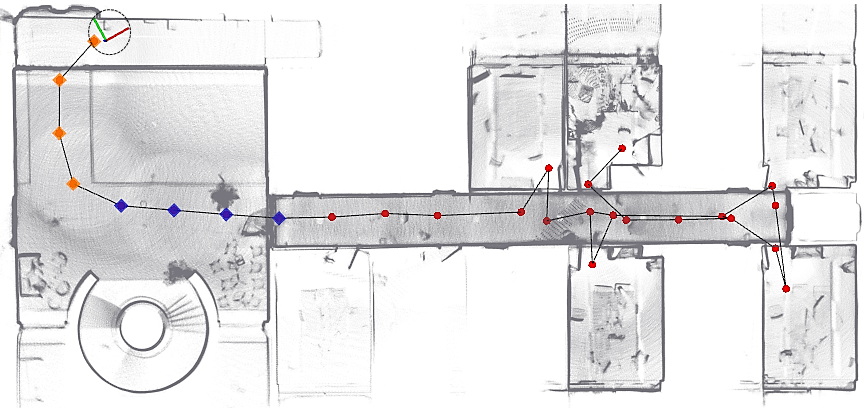}
  \caption{\textbf{Local Optimization.} Orange- and blue-colored keyframes, along with their connected layers, are included in the local optimization, with blue keyframes being fixed. Red keyframes are not incorporated in the optimization, as they are outside the optimization window.}
  \label{fig:local_optimization}
\end{figure}
\subsection{Local Optimization}
As a robot navigates through a scene we generate specific keyframe nodes and its observations thus incorporating new nodes and its edges within the graph. Each time the graph expands, we perform local optimization on a subset of the complete graph called the local graph, which consists of a sliding window containing the most recent $^M\boldsymbol{K}_n$ keyframes. In order to incorporate nodes and edges belonging to other hierarchical layers of the graph into local graph, we first check the connections between the $^M\boldsymbol{K}_n$ keyframes and the wall layer, incorporating all connected wall nodes $^M\boldsymbol{\Pi}_p$ and their corresponding edges. For all wall nodes $^M\boldsymbol{\Pi}_p$ within the optimization window, we check whether the entire set or a subset belongs to two-wall rooms $^M\boldsymbol{\rho}_s$ or four-wall rooms $^M\boldsymbol{\kappa}_k$. If such connections exist, we incorporate these room nodes into the local optimization.

\noindent Furthermore, we incorporate the corresponding floor node $^M\boldsymbol{\xi}_f$ that the robot is currently navigating into the local graph.
% In order to maintain consistency of the graph during optimization, we follow the strategy proposed in \cite{orb_slam3} to fix keyframes that are outside the current local optimization window but observe any of the walls $\boldsymbol{\pi}_p$ inside the local window.
To maintain graph consistency during optimization, we adopt the strategy proposed in \cite{orb_slam3}, fixing keyframes that lie outside the current optimization window but observe walls $^M\boldsymbol{\Pi}_p$ within the local window.
The state $\boldsymbol{s}_l$ to be optimized is defined as: 
\begin{equation}
    \boldsymbol{\hat{s}}_l = \argmin_{\boldsymbol{s}_l} (c_{\boldsymbol{k}_n}, c_{\boldsymbol{k_g}}, c_{\boldsymbol{\pi}_p}, c_{\boldsymbol{\rho_s}}, c_{\boldsymbol{\xi}_f})
\end{equation}
where $c_{\boldsymbol{k}_n}, c_{\boldsymbol{\pi}_p}, c_{\boldsymbol{\rho_s}}$ are the cost functions for the $n, p, s$  unfixed keyframes, walls and rooms respectively , ${c_{k_g}}$ is the cost of the $g$ fixed keyframes and $ c_{\boldsymbol{\xi}_f}$ is the cost of the $f^{th}$ floor node currently being explored. 
Fig.~\ref{fig:local_optimization} shows the local optimization strategy, with orange color keyframes along with the connected layers being optimized while the fixed keyframes are highlighted in blue. 
\begin{figure}[h]
  \centering
  \includegraphics[width=0.55\columnwidth]{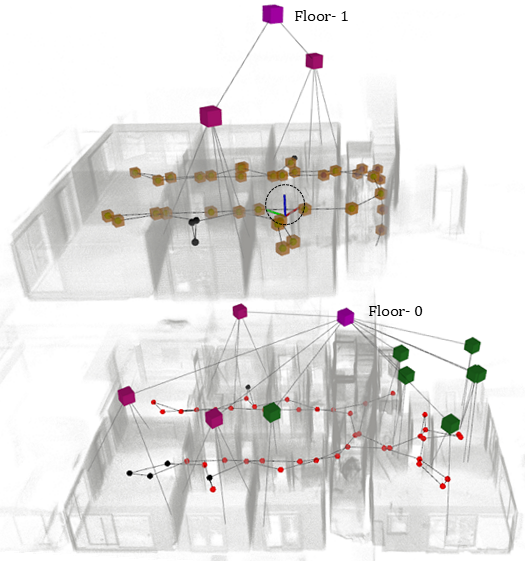}
  \caption{\textbf{Floor-level Global Optimization.} Floor-1 keyframes with orange colored boxes along with the connected wall, room and floor nodes are included in the floor-level global optimization on detection of a loop closure candidate on floor-1.  Floor-0 (lower level) maintains its own independent pose graph structure with keyframes (orange spheres) with connected rooms (pink and green cubes) and walls. These elements are temporarily excluded from the Floor-1 optimization to reduce computational complexity and prevent cross-floor interference.}
  \label{fig:floor_global_optimization}
\end{figure}
\subsection{Floor-level Global Optimization}
Floor-level global optimization is performed in mainly two situations, specifically, when a loop closure is found between the keyframes of a floor-level, and when a duplicate wall is found when a room is re-detected as explained in \cite{s_graphs+}. Our main motivation for this strategy is as follows.  When a new floor-level is explored, and a loop closure is found between two keyframes of such floor-level, this loop closure does not affect the keyframes or other connected layers of the preceding floor-levels. The same reasoning applies when duplicate walls are found for a detected room. 
When executing floor-level global optimization for floor-level $f$, we create a subset of the graph containing floor node $\boldsymbol{\xi}_f$, keyframe nodes $\boldsymbol{K}_{{n}_{\xi_f}}$ and its edges with the neighboring keyframes, then incorporate all wall nodes $\boldsymbol{\pi}_{p_{\xi_f}}$ with their edges to keyframe nodes $\boldsymbol{K}_{{n}_{\xi_f}}$. Floor-level global optimization incorporates all previously identified loop closure (relative pose) edges between the keyframes of that floor-level. 
We also incorporate four-wall $\boldsymbol{\rho}_{n_{\xi_f}}$ and two-wall room nodes $\boldsymbol{\kappa}_{n_{\xi_f}}$ with their edges to the underlying walls, finally also incorporating the edges between these rooms and the current floor node. To maintain consistency of the graph during optimization, we fix a keyframe which can either be the initial keyframe when the robot starts navigating, or it can be the last keyframe from the previous visited floor-level, this assures that the current floor-level map estimates do not diverge away from the previous floor-level map (See Fig.~\ref{fig:floor_global_optimization}). The state $\boldsymbol{s}_g$ using floor-level global optimization is optimized as:
\begin{multline}
\boldsymbol{\hat{s}}_g = \argmin_{\boldsymbol{s}_g} (c_{\boldsymbol{K}_{{n}_{\xi_f}}}, c_{\boldsymbol{K}_{{g}_{\xi_f}}}, c_{\boldsymbol{K}_{{l}_{\xi_f}}}, \\ 
c_{\boldsymbol{\pi}_{p_{\xi_f}}}, c_{\boldsymbol{\rho}_{s_{\xi_f}}}, c_{\boldsymbol{\kappa}_{k_{\xi_f}}}, c_{\boldsymbol{\xi}_f})
\end{multline}
where $c_{\boldsymbol{K}_{{n}_{\xi_f}}}, c_{\boldsymbol{K}_{{g}_{\xi_f}}}, c_{\boldsymbol{K}_{{lc}_{\xi_f}}}$ are the cost functions for the $n$ unfixed, $g$ fixed, $l$ loop closure keyframes. $c_{\boldsymbol{\pi}_{p_{\xi_f}}}, c_{\boldsymbol{\rho}_{s_{\xi_f}}}, c_{\boldsymbol{\kappa}_{k_{\xi_f}}}$ are the costs of $p$, $s$ and $k$ walls, two-wall and four-wall rooms respectively. $c_{\boldsymbol{\xi}_f}$ is the cost of the current floor node being explored. 

As a relevant effect of floor-level global optimization, if a robot \textit{revisits} a given floor-level and finds a suitable loop closure candidate, the floor-level global optimization incorporates all the nodes and edges not only from the current floor-level but also from all the previous visited ones. This approach also enables loop closures spanning multiple floors through staircases, ensuring comprehensive error correction across the entire multi-floor structure. This ensures the correction of all the accumulated errors from the revisited floor-level, as well as the previously visited ones.   
\subsection{Room-level Local Optimization}
Room-level local optimization is performed every time a four-wall room (bounded by four walls) is detected by the room segmentation module and room keyframes belonging to that particular room are identified (Section.~\ref{sec:room_keyframe_segmentation}). The room-level local optimization creates a graph subset containing the room node $\boldsymbol{\rho}_i$ along with the walls nodes $\boldsymbol{\pi}_{p_{\rho_i}}$ lying within the room along with the connected keyframes $\boldsymbol{K}_{n_{\rho_i}}$. Additionally as in local and floor-level global optimization to maintain the consistency of the graph we incorporate fixed set of keyframes $\boldsymbol{K}_{g_{\rho_i}}$ which are outside the room but observe the walls of the room. The state $\boldsymbol{s}_r$ during room-level local optimization is given as:
\begin{equation}
    \boldsymbol{\hat{s}}_r = \argmin_{\boldsymbol{s}_r} (c_{\boldsymbol{k}_{n_{\rho_i}}}, c_{\boldsymbol{k}_{g_{\rho_i}}}, c_{\boldsymbol{\pi}_{p_{\rho_i}}}, c_{\boldsymbol{\rho_i}})
\end{equation}
Where $c_{\boldsymbol{k}_{n_{\rho_i}}}, c_{\boldsymbol{k}_{g_{\rho_i}}}$ is the cost for $n$ unfixed and $g$ fixed keyframes and $c_{\boldsymbol{\pi}_{p_{\rho_i}}}$ and $c_{\boldsymbol{\rho_i}}$ are the costs related to $p$ walls and the $i$-th room. 

\begin{figure}[t]
  \centering
  \includegraphics[width=0.8\columnwidth]{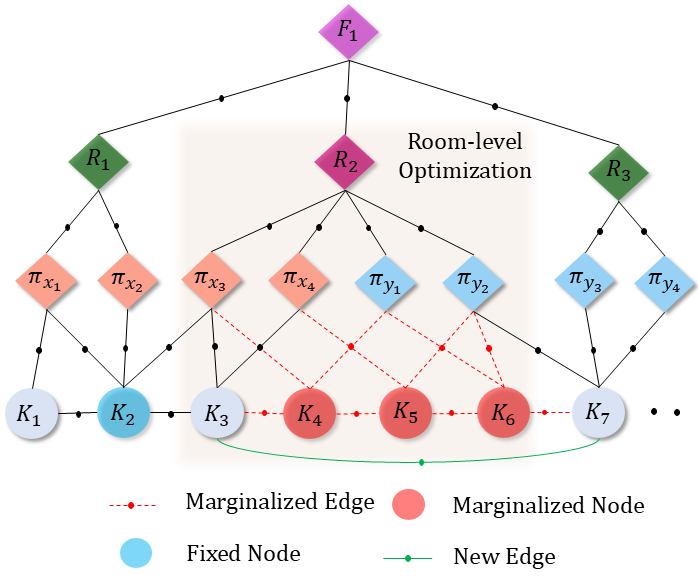}
  \caption{\textbf{Room-level Local Optimization.} Overview of the compressed graph after room-level local optimization. Orange box shows nodes used for room-level optimization, while the dotted red box indicates keyframe marginalization post-optimization. The green line shows the new edge connecting the first keyframe of the room with the first keyframe outside the room.}
  \label{fig:room_local_optimization}
\end{figure}

The main strategy of the room-level local optimization is to marginalize redundant keyframe nodes and its edges that observe the same room structure. Thus, after a room-level local optimization is performed, all the keyframe nodes and it corresponding edges except the first keyframe node/edges are marginalized out from the main global graph. 
%The marginalized information from the room-level local optimization step is included to generate a compressed global graph excluding the marginalized keyframes and their edges. 
Marginalization leads to a generation of a disconnected graph between keyframes. These disconnections are between the marginalized keyframes and their non-marginalized neighbors. To obtain a connected global graph, we incrementally check for all the edges $e_{i,n}$, between the $i$-th marginalized keyframe $\boldsymbol{K}_i$ with $n$-th non-marginalized neighbor $\boldsymbol{K}_n$, and connect the closest non-marginalized neighbors $\boldsymbol{K}_{i-1}$ and $\boldsymbol{K}_n$ with a new edge $e_f$ (See Fig.~\ref{fig:room_local_optimization}). The information matrix of the new edge is the summation of the information matrices of the edges removed in $e_{i,n}$:
\begin{equation}
    \Omega_{e_f} = \sum_{(\boldsymbol{K}_i, \boldsymbol{K}_n) \in \mathcal{E}_m} \Omega_{e_{i,n}}
\end{equation}
where $\Omega_{e_{i,n}}$ is the information matrix of the original edge $e_{i,n}$ and $\Omega_{e_f}$ is the resulting information matrix for the new edge $e_f = (K_{i-1}, K_n)$.
The new edge $e_f$ is added back to the global graph edges:
\begin{equation}
    \mathcal{E}_{global} \leftarrow \mathcal{E}_{global} \cup \{ (K_{i-1}, K_n, \Omega_{e_f}) \}
\end{equation}
The room-level local optimization is always performed after the execution of at least one of the local or floor-level global optimization.

%% file: experimental_results.tex
\section{Experimental Results}
\subsection{Methodology}
We validate our algorithm on different real indoor environments (single-floor and multi-floor) from university buildings to construction sites, validating it with different state-of-the-art SLAM algorithms, including the baseline \textit{S-Graphs+} \cite{s_graphs+}. Our datasets are recorded using Velodyne VLP-16 or Ouster OS-1 64 3D LiDAR, with odometry from robot encoders or FAST-LIO2 \cite{fast_lio2}. These sensors demonstrate that our algorithm is sensor-agnostic. Despite differing beam counts, both sensors are downsampled to uniform resolution for processing. Dynamic objects such as humans are present in many of these datasets; while handling dynamic objects falls outside the scope of this paper, the reported results reflect their presence as we neither explicitly include nor exclude them from our analysis.

\textbf{Single-Floor Dataset.} 
The single-floor dataset consists of small-scale indoor environments with the robot navigating around one single floor. The main motivation for these experiments is to fairly validate the accuracy and cost reduction of our S-Graphs 2.0 against baselines, as most baselines accumulate significant drift or incur in false loop closures for multi-floor scenes. The first two experiments, \textit{C1F1} and \textit{C1F2}, are performed on two different floors of a construction site. Additionally, \textit{C2F0}, \textit{C2F1}, and \textit{C2F2} consist of three different floors of an ongoing construction site combining four individual houses. \textit{C3F1}, and \textit{C3F2} are two different floors two combined houses. For validation, due to the absence of ground truth trajectory metrics, we use ground truth point clouds, extracted from available Building Information Models (BIM), to report pointcloud RMSE. Furthermore, we also report the computational time of our approach against the most accurate baseline \textit{S-Graphs+} \cite{s_graphs+}. 

\textbf{Multi-Floor Dataset.}
The multi-floor dataset is aimed at validating the accuracy of our algorithm in the presence of larger trajectories covering multiple floors. It consists of five datasets, \textit{BE1} and \textit{BE2} are university buildings where the robot navigates the three and two floors, respectively. \textit{LC1} is a university building with long corridors and a robot traversing two floors. \textit{CS1} and \textit{CS2} are buildings under construction with a robot traversing all three floors.
Due to the absence of ground truth, to validate the accuracy of each approach, we use the Mean Map Entropy (MME) presented in \cite{efficient_continous_SLAM} to measure the sharpness of the generated map.
Additionally, to assess algorithmic accuracy in multi-floor segmentation, we evaluate the Intersection over Union (IoU) between the histogram of z-value occurrences per floor level in the generated map and the percentage of ground truth time spent per floor level. In this evaluation, we recorded the time the robot spent on each floor level, which was subsequently converted into percentage data for that floor. We also documented the maximum and minimum heights for each floor level, establishing the ground truth bins for the datasets. We then utilized the z-values from the 3D maps generated by our algorithms and baselines to compute the IoU of these bins against the ground truth bins. While IoU is an indirect metric, it captures if the SLAM algorithm correctly tracks the robot's position. The vertical distribution of mapped points should match the time spent on each floor. When the algorithm incorrectly identifies floors, points are mapped at incorrect heights, resulting in clear mismatches in the z-distribution. Thus, high IoU indicates good floor-level tracking throughout the mapping process, providing meaningful validation despite the lack of ground truth poses.
\subsection{Results and Discussions}
\textbf{Single-Floor Dataset.}
The results of the single-floor experiments, in Tab.~\ref{tab:rmse_real_data}, show that our \textit{S-Graphs 2.0} is able to provide final map accuracy similar to the second best baseline \textit{S-Graphs+} with an average accuracy improvement of $0.48\%$. However, the benefits of the hierarchical optimization of our \textit{S-Graphs 2.0} clearly stand out in Tab.~\ref{tab:computation_time}, where it has an average cost reduction of $79.87\%$ to the closest baseline. The benefits of the hierarchical optimization become even more clear for longer sequences. See, for example, how in \textit{C2F2}, \textit{C3F2} there is slight decrease in the accuracy with respect to \textit{S-Graphs+} ($2.27\%$ and $2.67\%$ respectively),but a large reduction the computation time ($85.56\%$ and $82.66\%$ respectively). Summing up, in these single-floor datasets, we show that the optimization strategy in \textit{S-Graphs 2.0} maintains the estimation accuracy while significantly reducing the computational cost. 
We also compared our algorithm and other baselines on the publicly available Hilti SLAM datasets \cite{hilti_dataset}. For fair comparison, we use the same odometry source \emph{fast lio} \cite{fast_lio2} for all. Tab. \ref{tab:RMSE_hilti_data} reports the ATE for the easy (E04, E05) and medium (E06, E14) difficulty categories. We used  Our \textit{S-Graphs 2.0} consistently outperforms the baseline in all sequences. Note that none of the methods converged on the hard category sequences due to their extreme difficulty.

\begin{table}[h]
\centering
\renewcommand{\arraystretch}{0.8} 
\setlength{\tabcolsep}{4pt}       
\scriptsize
\caption{Point cloud RMSE [cm] for our single-floor dataset. Best results are \textbf{boldfaced}, second best are \underline{underlined}.}
\begin{tabular}{l|ccccccc|c}
\toprule
\multirow{2}{*}{\textbf{Method}} & \multicolumn{7}{c|}{\textbf{Point Cloud RMSE [cm] $\downarrow$}} & \multirow{2}{*}{\textbf{Avg}} \\
\cmidrule(lr){2-8}
 & \textit{C1F1} & \textit{C1F2} & \textit{C2F0} & \textit{C2F1} & \textit{C2F2} & \textit{C3F1} & \textit{C3F2} \\
\midrule
HDL-SLAM \cite{hdl_graph_slam} & 33.5 & 19.8 & 18.5 & 21.1 & 19.5 & 22.9 & 19.4 & 22.1 \\
ALOAM \cite{loam}             & 52.6 & 33.6 & 34.1 & 45.1 & 29.9 & 36.5 & 43.4 & 39.3 \\
SCA-LOAM \cite{scan_context}   & 45.6 & 26.1 & 22.5 & 25.4 & 20.2 & 26.9 & 19.2 & 26.6 \\
BALM \cite{BALM}              & 34.9 & 22.5 & 18.2 & 19.7 & 99.0 & 26.9 & 22.6 & 34.8 \\
S-Graphs+ \cite{s_graphs+}     & \underline{32.9} & \textbf{18.9} & \textbf{16.9} & \textbf{18.9} & \textbf{17.6} & 22.3 & \textbf{18.7} & \underline{20.9} \\
\midrule
\textit{Ours}                & \textbf{31.3} & \underline{19.0} & \underline{17.0} & \underline{19.3} & \underline{18.0} & \textbf{22.0} & \underline{19.2} & \textbf{20.8} \\
\bottomrule
\end{tabular}
\label{tab:rmse_real_data}
\end{table}
\begin{table}[h]
\centering
\renewcommand{\arraystretch}{0.8} 
\setlength{\tabcolsep}{4pt}     
\scriptsize
\caption{Absolute Trajectory Error (ATE) in \textit{cm} for Hilti datasets \cite{hilti_dataset}. Best results are \textbf{boldfaced}, second best are \underline{underlined}.}
\begin{tabular}{l|cccc|c}
\toprule
\multirow{2}{*}{\textbf{Method}} & \multicolumn{4}{c|}{\textbf{Absolute Trajectory Error $\downarrow$}} & \multirow{2}{*}{\textbf{Avg}} \\
\cmidrule(lr){2-5}
 & \textit{E04} & \textit{E05} & \textit{E06} & \textit{E14} &  \\
\midrule
 HDL-SLAM \cite{hdl_graph_slam} & 21.9 & 15.4 & \underline{120.4} & 14.2 & 42.9  \\ 
ALOAM \cite{loam} &  28.2 & 22.5 &  130.1 &  24.9 &  51.4  \\ 
SCA-LOAM \cite{scan_context}   &  28.6 &  23.1 &  130.4 &  25.3 &  51.9 \\ 
 BALM \cite{BALM}  & 27.9 & 18.8 &  129.4 &  16.7 &  48.2 \\ 
S-Graphs+ \cite{s_graphs+}     & \underline{20.5}  & \underline{13.8}  & {123.2}  &  \underline{13.1} &  \underline{42.6} \\ 
\midrule
\textit{Ours}                & \textbf{19.8} & \textbf{13.0} & \textbf{50.3} & \textbf{11.5} & \textbf{23.6} \\
\bottomrule
\end{tabular}
\label{tab:RMSE_hilti_data}
\end{table}

\begin{table}[h]
\centering
\renewcommand{\arraystretch}{0.8} 
\setlength{\tabcolsep}{4pt}     
\scriptsize
\caption{Point cloud Mean Map Entropy (MME) for our multi floor dataset. Best results are \textbf{boldfaced}, second best are \underline{underlined}.}
\begin{tabular}{l|ccccc|c}
\toprule
\multirow{2}{*}{\textbf{Method}} & \multicolumn{5}{c|}{\textbf{Point Cloud MME $\downarrow$}} & \multirow{2}{*}{\textbf{Avg}} \\
\cmidrule(lr){2-6}
 & \textit{BE1} & \textit{BE2} & \textit{LC1} & \textit{CS1} & \textit{CS2} & \\
\midrule
HDL-SLAM \cite{hdl_graph_slam} & -1.23 & \underline{-1.38} & -1.31 & -0.97 & \underline{-1.14} & -1.21 \\ 
ALOAM \cite{loam}             & \underline{-1.37} & -1.23 & -1.32 & -0.73 & -1.07 & -1.14 \\ 
SCA-LOAM \cite{scan_context}   & -0.85 & -1.04 & -0.96 & -0.66 & -1.04 & -0.91 \\ 
BALM \cite{BALM}              & -1.35 & -1.27 & \underline{-1.62}  & -0.77 & \textbf{-1.28} & \underline{-1.26} \\ 
S-Graphs+ \cite{s_graphs+}     & -1.23 & -1.28 & -1.37 & \underline{-1.06} & -1.11 & -1.21 \\ 
\midrule
\textit{Ours}                & \textbf{-1.58} & \textbf{-1.41} & \textbf{-1.67} & \textbf{-1.24} & \textbf{-1.28} & \textbf{-1.44} \\
\bottomrule
\end{tabular}
\label{tab:mme_real_data}
\end{table}

\begin{table}[h]
\centering
\renewcommand{\arraystretch}{0.8} 
\setlength{\tabcolsep}{4pt}     
\scriptsize
\caption{Intersection over Union (IoU) for our multi-floor dataset. Best results are \textbf{boldfaced}, second best \underline{underlined}.}
\label{tab:iou_multi_floor}
\begin{tabular}{l|ccccc|c}
\toprule
\multirow{2}{*}{\textbf{Method}} & \multicolumn{5}{c|}{\textbf{IoU} $\uparrow$} & \multirow{2}{*}{\textbf{Avg}} \\
\cmidrule(lr){2-6}
 & \textit{BE1} & \textit{BE2} & \textit{LC1} & \textit{CS1} & \textit{CS2} & \\
\midrule
HDL-SLAM \cite{hdl_graph_slam} & 0.76 & \underline{0.90} & 0.55 & 0.37 & 0.35 & 0.59 \\ 
ALOAM \cite{loam}             & 0.50 & 0.78 & 0.69 & 0.69 & 0.31 & 0.59 \\ 
SCA-LOAM \cite{scan_context}   & 0.56 & 0.81 & \underline{0.78} & 0.51 & 0.31 & 0.59 \\ 
BALM \cite{BALM}              & 0.67 & 0.71 & 0.77 & \underline{0.79} & \underline{0.80} & \underline{0.75} \\ 
S-Graphs+ \cite{s_graphs+}     & \underline{0.83} & 0.72 & 0.46 & 0.33 & 0.30 & 0.53 \\ 
\midrule
\textit{Ours}  & \textbf{0.90} & \textbf{0.91} & \textbf{0.93} & \textbf{0.90} & \textbf{0.91} & \textbf{0.91} \\
\bottomrule
\end{tabular}
\end{table}

\begin{figure}[h]
  \centering
  \includegraphics[width=0.75\columnwidth]{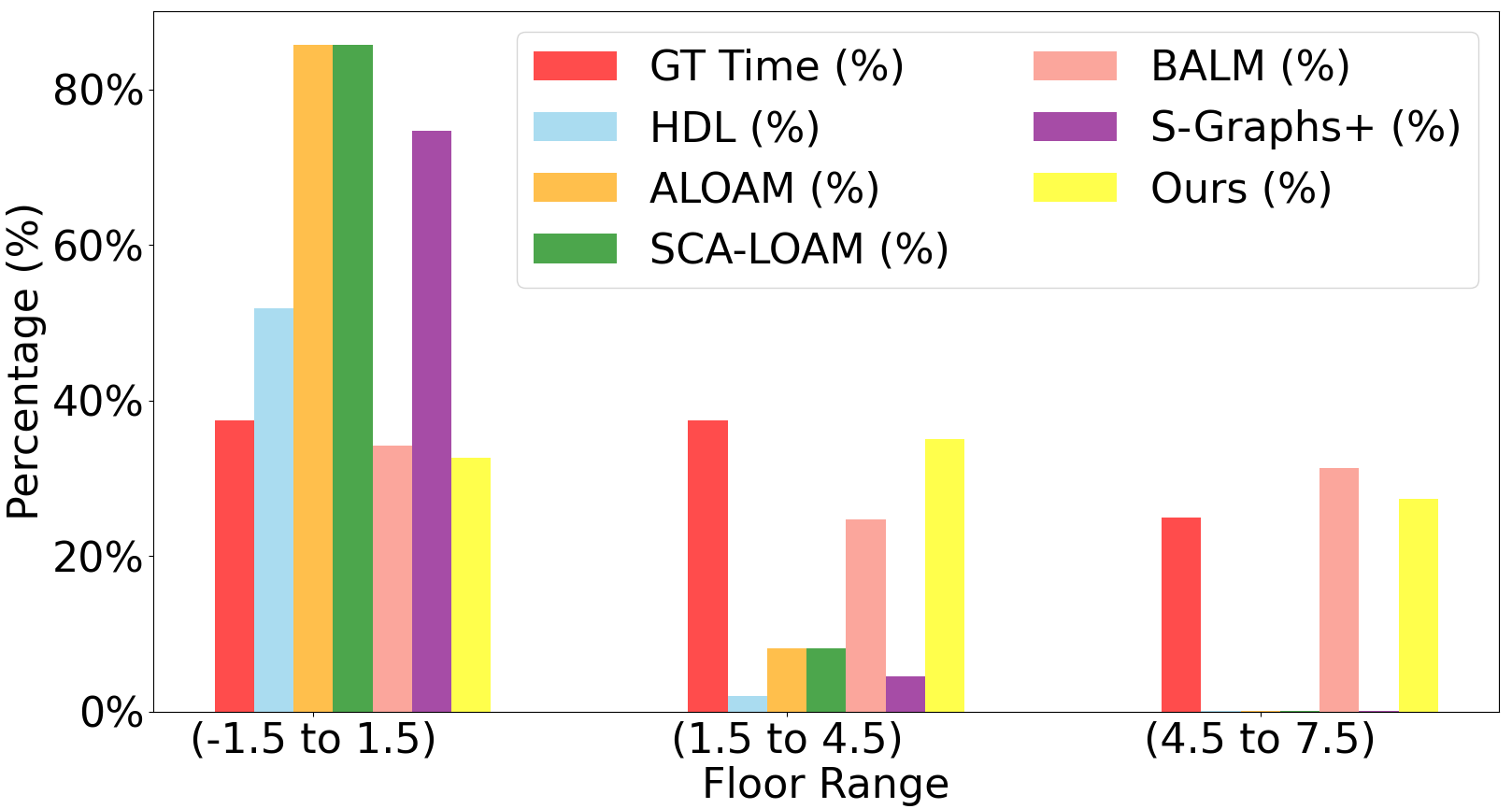}
  \caption{\textbf{Floor Comparison.} \textit{CS2} dataset comparing the distribution of time spent per floor (ground truth) in percentage with the distribution of map-derived z-values converted to percentages, organized per floor.}
  \label{fig:iou}
\end{figure}

\begin{table}[h]
\centering
\renewcommand{\arraystretch}{0.7} % reduce row height
\setlength{\tabcolsep}{3pt}       % adjust column spacing
% \footnotesize
\scriptsize
\caption{Computation time in milliseconds (ms) for our method compared against the baseline \textit{S-Graphs+}. Sequence lengths [s] are indicated for each dataset.}
\label{tab:computation_time}
\begin{tabular}{l c | c c}
\toprule
\textbf{Dataset} & \textbf{Seq. Length [s]} & \multicolumn{2}{c}{\textbf{Computation Time [ms]}} \\
\midrule & & S-Graphs+ & \textit{Ours} \\
\midrule
{\textit{C1F1}} & 487   & 74.0  & \textbf{36.0} \\
{\textit{C1F2}} & 657   & 106   & \textbf{38.0} \\
{\textit{C2F0}} & 238   & 87.3  & \textbf{4.00} \\
{\textit{C2F1}} & 672   & 169   & \textbf{23.0} \\
{\textit{C2F2}} & 1044  & 263   & \textbf{37.0} \\
{\textit{C3F1}} & 558   & 125   & \textbf{8.00} \\
{\textit{C3F2}} & 999   & 173   & \textbf{30.0} \\
\midrule
{\textit{BE1}}  & 1378  & 1106  & \textbf{88.0} \\
{\textit{BE2}}  & 1032  & 479   & \textbf{101}  \\
{\textit{LC1}}  & 490   & 180   & \textbf{21.0} \\
{\textit{CS1}}  & 3000  & 1085  & \textbf{10.0} \\
{\textit{CS2}}  & 690   & 126   & \textbf{13.0} \\
\midrule
\textbf{Avg} & --  & 331.1 & \textbf{34.08} \\
\bottomrule
\end{tabular}
\end{table}

\begin{figure}[t]
\centering
\begin{subfigure}[h]{.21\textwidth}
\centering
\includegraphics[width=1\textwidth]{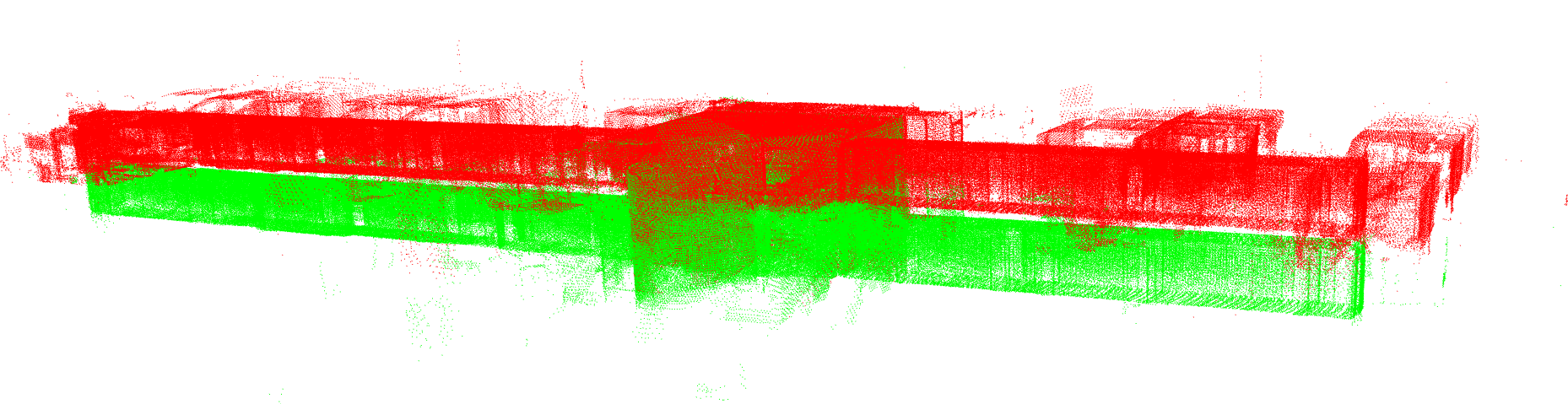}
\caption{{\textit{S-Graphs 2.0 (LC1)}}}
\label{fig:3d_map_s_graphs}
\end{subfigure}
\begin{subfigure}[h]{0.21\textwidth}
\centering
\includegraphics[width=1\textwidth]{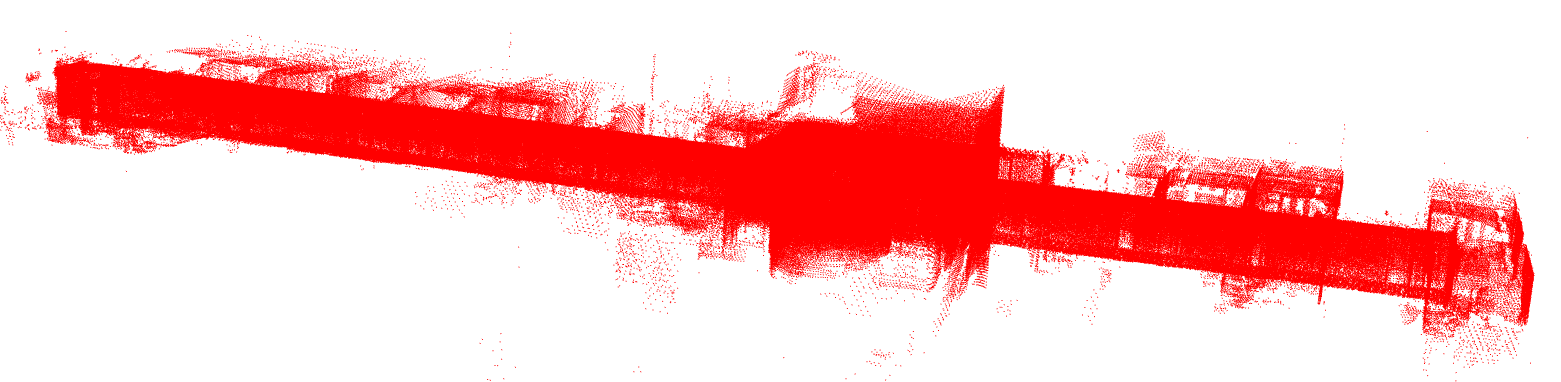}
\caption{\textit{S-Graphs+ (LC1)}}
\label{fig:3d_map_hdl_slam}
\end{subfigure}
\begin{subfigure}[h]{.21\textwidth}
\centering
\includegraphics[width=1\textwidth]{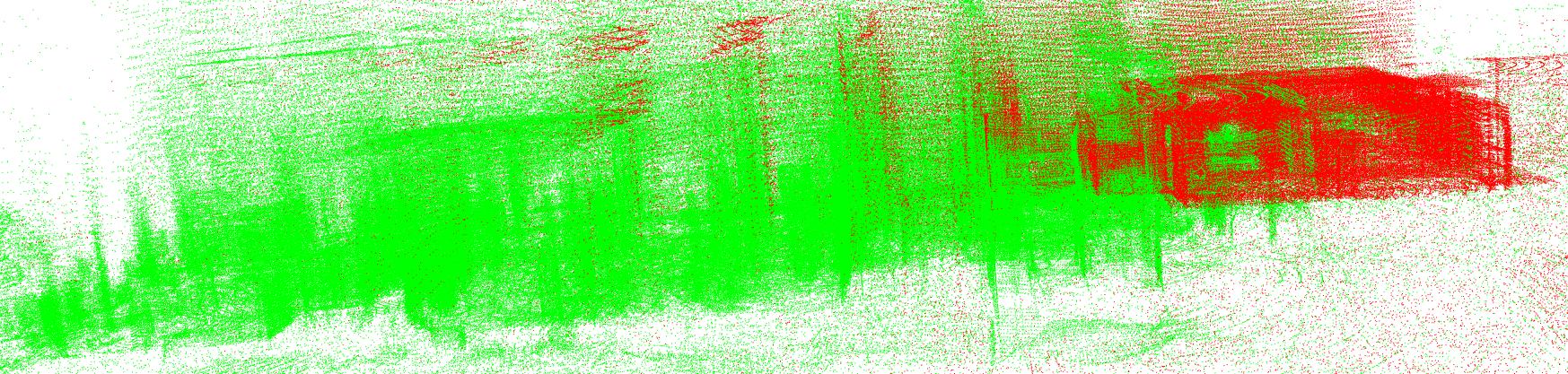}
\caption{{\textit{S-Graphs 2.0 (Hilti)}}}
\label{fig:map_s_graphs_2_hilti}
\end{subfigure}
\begin{subfigure}[h]{0.21\textwidth}
\centering
\includegraphics[width=1\textwidth]{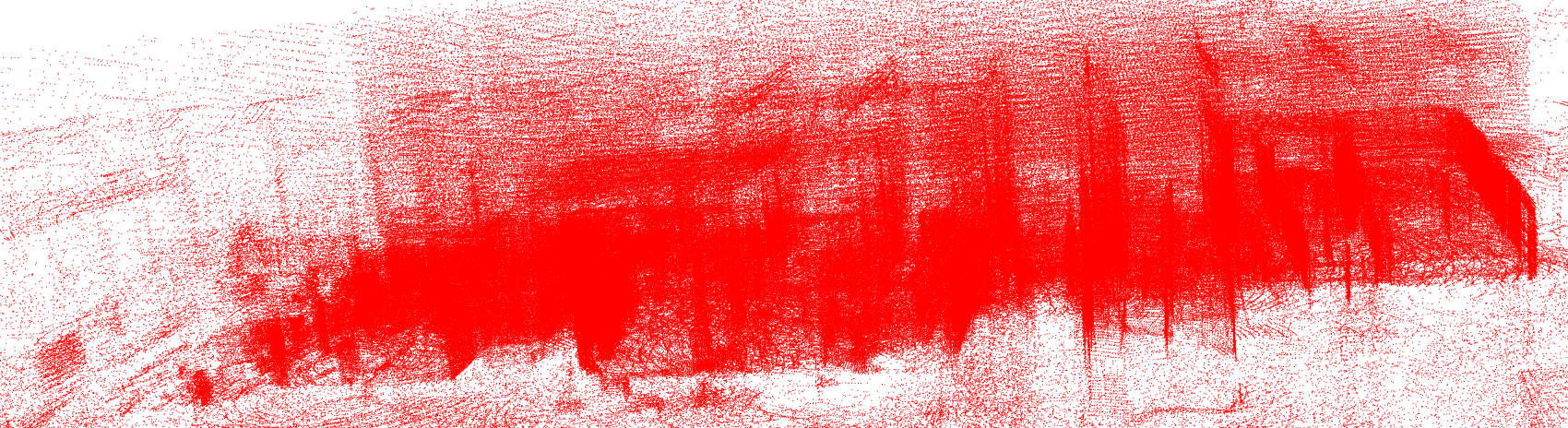}
\caption{\textit{S-Graphs+ (Hilti)}}
\label{fig:map_s_graphs_p_hilti}
\end{subfigure}
\caption{\textbf{Qualitative Results} comparing \textit{S-Graphs 2.0} and \textit{S-Graphs+} Top row (a-b): multi-floor dataset \textit{LC1}, bottom row (c-d): \textit{Hilti} dataset\cite{hilti_dataset} . Note how the two floor levels collapse for \textit{S-Graphs+}, while our \textit{S-Graphs 2.0} shows correct results.}
\label{fig:3d_map_l1c1}
%\vspace{-2mm}, while our 
\end{figure}

\begin{figure}[!h]
\centering
\begin{subfigure}[t]{.15\textwidth}
\centering
\includegraphics[width=1\textwidth]{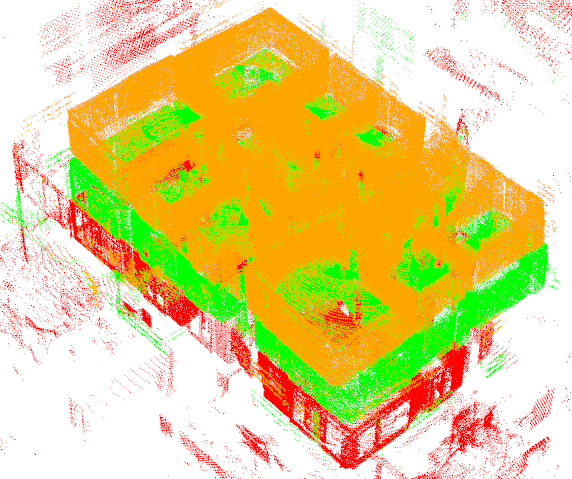}
\caption{{\textit{S-Graphs 2.0}}}
\label{fig:3d_map_s_graphs}
\end{subfigure}
\begin{subfigure}[t]{0.15\textwidth}
\centering
\includegraphics[width=1\textwidth]{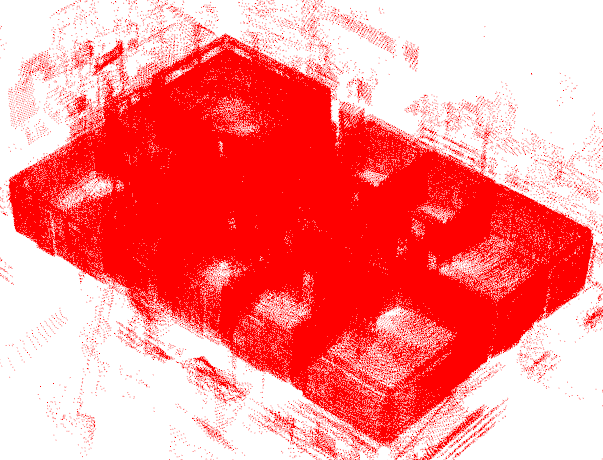}
\caption{\textit{S-Graphs+}}
\label{fig:3d_map_hdl_slam}
\end{subfigure}
\begin{subfigure}[t]{0.15\textwidth}
\centering
\includegraphics[width=1\textwidth]{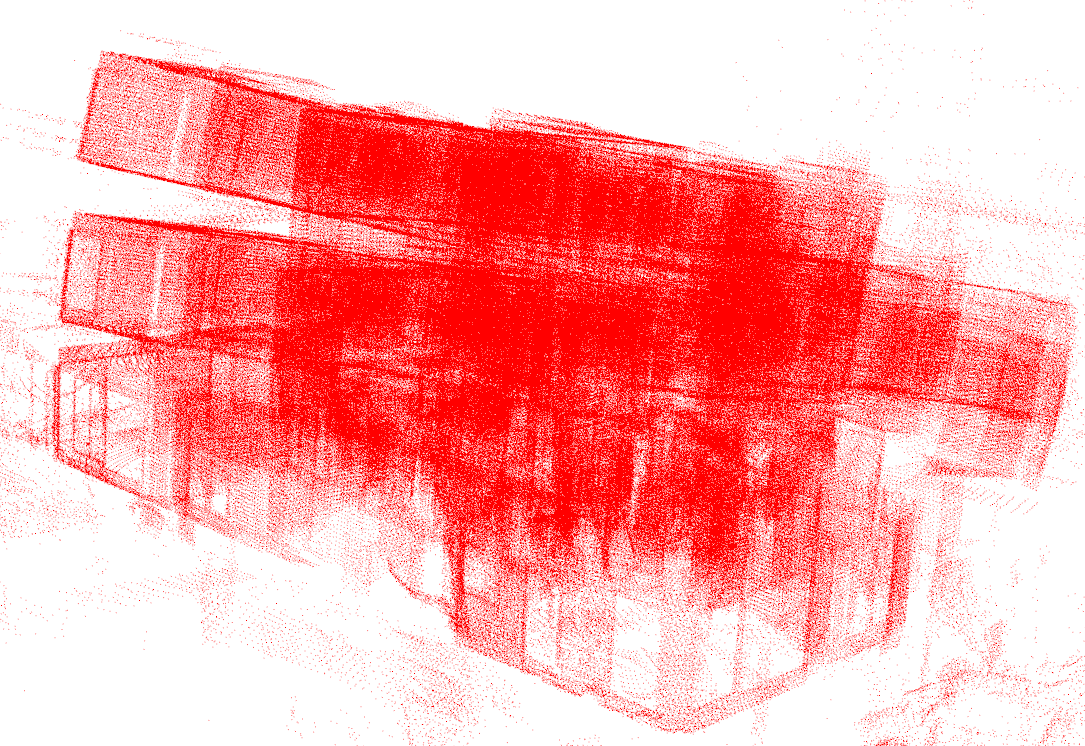}
\caption{\textit{BALM}}
\label{fig:3d_map_hdl_slam}
\end{subfigure}
\caption{\textbf{Qualitative Results} comparing \textit{S-Graphs 2.0}, \textit{S-Graphs+} and \textit{BALM} in the multi-floor dataset \textit{CS2}. Note the collapse of floor levels for \textit{S-Graphs+}, deviations for \textit{BALM} and accurate estimation results for our \textit{S-Graphs 2.0}.}
\label{fig:3d_map_CS2}
%\vspace{-2mm}
\end{figure}

\textbf{Multi-Floor Dataset.}
Tab.~\ref{tab:mme_real_data} shows the Mean Map Entropy (MME) for the multi-floor dataset. Our method outperforms all the baselines by an average of $25.67\%$ when comparing the MME. Given the large area covered in these multi-floor experiments, all baselines fail at giving reasonable estimates, being unable to maintain the separation between the floor levels as well as accumulating significant drift. This can be assessed in  Fig.~\ref{fig:3d_map_l1c1}, which shows the maps estimated by \textit{S-Graphs 2.0} and \textit{S-Graphs+} for \textit{LC1} (Top row (a-b) respectively), and in one of the multifloor Hilti datasets \cite{hilti_dataset}, bottom row (c-d). As these environments are highly aliased between floors, \textit{S-Graphs+}'s loop closure gives a high number of false positives between levels. Our approach, given the floor segmentation, stairway detection, and floor-based loop closure, is able to maintain the accurate map while segmenting the keyframes and their connected layers based on the given floor-level. Fig.~\ref{fig:3d_map_CS2} shows qualitative results for \textit{CS2} demonstrating that our method is capable of segmenting the three floor-levels while maintaining a good map accuracy compared to \textit{S-Graphs+} (which again performs inaccurate loop closures between floor levels) and BALM. Tab.~\ref{tab:iou_multi_floor} and Fig.~\ref{fig:iou} also validate the proper floor segmentation by comparing the IoU percentage of $z$ (height) values per floor-level from the maps generated by the algorithms versus the percentage ground truth time spend per floor-level. Observe that the IoU of our \textit{S-Graphs 2.0} correlates perfectly with the ground truth time, \textit{BALM} presents a weaker correlation and the rest of baselines offer very poor results.  

Remarkably, as reported in Tab.~\ref{tab:computation_time}, the average reduction of computation time in multi-floor dataset of \textit{S-Graphs 2.0} compared to \textit{S-Graphs+} is $92.2\%$. As detailed before, this comes as a result of the hierarchical optimization and floor-level loop closure strategies of our \textit{S-Graphs 2.0}.

%% file: conclusion.tex
\section{Conclusion}

In this work, we present \textit{S-Graphs 2.0}, a multi-layered hierarchical SLAM for large, multi-floor indoor environments. Our approach leverages semantic relationships and a robust floor detection module to assign floor-level information, enabling effective floor-level global and room-level local optimizations as well as reliable loop closures across visually similar areas. We demonstrate a $19.48\%$ increase in accuracy and a $92.2\%$ reduction in computation time compared to the second-best method. While our current floor detection relies on trajectory slope analysis and thus does not handle purely vertical transitions such as elevators, it effectively manages the more common case of stair-based floor transitions. Future work will integrate additional sensing modalities to handle elevator transitions, as well as further semantic-relational concepts to enhance floor-based loop closure and optimization.